\pdfoutput=1

\documentclass[11pt]{article}

\usepackage[]{EMNLP2022}

\usepackage{times}
\usepackage{latexsym}

\usepackage{amsmath}
\usepackage{graphicx}
\usepackage{caption}
\usepackage{booktabs}
\usepackage{pifont}
\usepackage{subfigure}
\usepackage{bbm}
\usepackage{amsfonts}
\usepackage{multirow}
\usepackage[T1]{fontenc}

\usepackage[utf8]{inputenc}

\usepackage{microtype}

\usepackage{inconsolata}

\input{mySymbol.sty}

\title{Retrieval Augmented Visual Question Answering with Outside Knowledge}

\author{Weizhe Lin\\
  Department of Engineering\\
  University of Cambridge \\
  United Kingdom \\
  \texttt{wl356@cam.ac.uk} \\\And
  Bill Byrne \\
  Department of Engineering\\
  University of Cambridge \\
  United Kingdom \\
  \texttt{bill.byrne@eng.cam.ac.uk} \\}

\begin{document}
\maketitle
\begin{abstract}
Outside-Knowledge Visual Question Answering (OK-VQA) is a challenging VQA task that requires retrieval of external knowledge to answer questions about images.
Recent OK-VQA systems use Dense Passage Retrieval (DPR) to retrieve documents from external knowledge bases, such as Wikipedia, but with DPR trained separately from answer generation, introducing a potential limit on the overall system performance.
Instead, we propose a joint training scheme which includes differentiable DPR integrated with answer generation so that the system can be trained in an end-to-end fashion.
Our experiments show that our scheme outperforms recent OK-VQA systems with strong DPR for retrieval.
We also introduce new diagnostic metrics to analyze how retrieval and generation interact.
The strong retrieval ability of our model significantly reduces the number of retrieved documents needed in training, yielding significant benefits in answer quality and computation required for training.

\end{abstract}

\section{Introduction}
\label{s:intro}
Visual Question Answering (VQA) is a challenging problem that lies at the intersection of Computer Vision, Natural Language Processing, and Information Retrieval.
The objective in VQA is to read an image and provide an answer to an accompanying question about the image content. 
Current approaches to VQA employ deep-learning-based systems to jointly understand images and text.

\begin{figure}[!htp]
    \centering
    \includegraphics[width=0.9\linewidth]{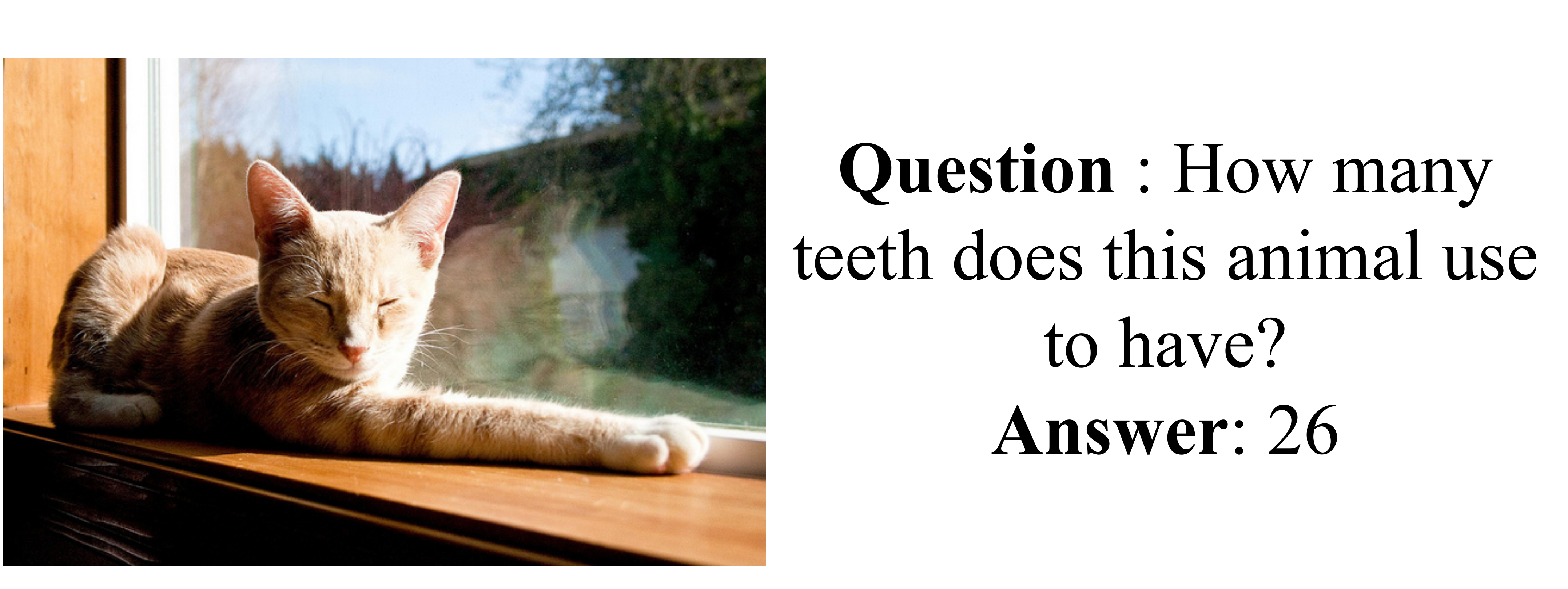}
    \caption{OK-VQA contains questions whose answer cannot be found within the image.}
    \label{fig:example}
\end{figure}

VQA is particularly challenging when the answer to the question is not directly available in the image.
In \textit{Knowledge-based} VQA (KB-VQA), the VQA system must access external knowledge sources to find a correct and complete answer.
The Ouside-Knowledge VQA task (OK-VQA)~\cite{marino2019ok}  consists of questions that requires general knowledge and simple inference to answer (Fig.~\ref{fig:example}).
Such questions are even hard for humans.
Unlike other KB-VQA datasets (e.g. FVQA~\cite{wang2017fvqa}) which provide an associated knowledge base, OK-VQA encourages using any outside knowledge in answering questions.

The need to adapt and refresh knowledge sources motivates the study of  KB-VQA systems that can extract knowledge from both structured  (e.g. ConceptNet~\cite{speer2017conceptnet}) and unstructured knowledge representations (e.g. Wikipedia passages).
Recent designs~\cite{luo-etal-2021-weakly, gao2022transform} approach VQA in two distinct steps: (1) \textit{Knowledge Retrieval} extracts documents from a large knowledge base; (2) \textit{Answer Generation} produces an answer from these documents.    Knowledge Retrieval can be done via 
Dense Passage Retrieval (DPR)~\cite{karpukhin-etal-2020-dense}, which 
consists of a question encoder and a document encoder (both Transformer-based) that encode questions and documents into separate dense representations.
The DPR system is trained to assign higher scores to documents intended to be helpful in answering questions, so that document sets can be retrieved and passed to Answer Generation.

Knowledge Retrieval based on DPR is powerful but has some readily observed limitations, particularly in model training.
Firstly, whether a retrieved document is useful in answering a question cannot be easily determined, even if an answer is provided.
Prior work \cite{qu2021passage, luo-etal-2021-weakly} has addressed this problem using ``\textit{Pseudo Relevance Labels}'' which are based on whether a document contains a given answer.
However, these are only a weak signal of potential relevance and may encourage DPR to retrieve misleading documents.
Secondly, the document retriever and answer generator are trained separately.   
To ensure that the answer generator sees relevant documents in training, systems can retrieve large numbers of documents  ($\sim$50+)~\cite{gao2022transform, gui2021kat}, but at the cost of slower training and more GPU usage, and also possibly presenting misleading material to the answer generator.

Joint training of the retriever and answer generator offers a solution to these problems.   The aim is twofold: (1) to improve the retrieval of documents truly relevant to providing a given answer; and (2) to reject documents with pseudo relevance but not actual relevance.

Retrieval Augmented Generation (RAG)~\cite{lewis2020retrieval} has shown that end-to-end joint training of a DPR-based QA system can outperform  baseline two-step systems.  A notable feature of  RAG is a loss function that incorporates marginalized likelihoods over retrieved documents such that the training score of a document is increased whenever it improves prediction.

However, in preliminary OK-VQA experiments we found that RAG did not perform well.
Our investigations found that a good portion of OK-VQA training questions are answerable in closed-book form (i.e. using pre-trained models such as T5~\cite{t5paper}) with information extracted only from the image, 
with the unintended consequence that the RAG loss function awards credit to documents that did not actually contribute to answering a question.
We also found that difficult questions that are unanswerable with the knowledge available to retrieval were more prevalent in OK-VQA than in the Open QA datasets (e.g. Natural Questions~\cite{kwiatkowski2019natural}) on which RAG was developed.   In both of these scenarios, the RAG loss function leads to counter-intuitive adjustments to the document scores used in training the retrieval model, leading to decreased VQA performance.

Motivated by these findings, we propose a novel neural-retrieval-in-the-loop framework for joint training of the retriever and the answer generator.  We formulate a loss function that avoids sending misleading signals to the retrieval model in the presence of irrelevant documents.     This formalism combines both pseudo relevance labels and model predictions to refine document scores in training.
We find significantly better performance on OK-VQA compared to RAG.
In this paper:
\begin{itemize}
\setlength\itemsep{-0.3em}
    \item We present a novel joint training framework \textbf{R}etrieval \textbf{A}ugmented \textbf{V}isual \textbf{Q}uestion \textbf{A}nswering (RA-VQA) for Knowledge Retrieval and Answer Generation that improves over RAG and two-step baseline systems based on DPR~\cite{karpukhin-etal-2020-dense}.
    \item We investigate  visually grounded features transformed into `language space' and assess their contribution to OK-VQA performance.
    \item We study the role of document retrieval in KB-VQA and evaluate its interaction with retrieval-augmented generation.
    We also show that retrieval becomes more efficient in joint training, requiring retrieval of relatively few ($\sim5$) documents in training. 
%
    
\end{itemize}

\section{Related Work}
\label{sec:related_work}
\textbf{Open-domain QA systems.} These QA systems are designed to answer questions from datasets such as Natural Questions~\cite{kwiatkowski2019natural}.
The knowledge needed to answer questions can be in pre-trained models~\cite{roberts-etal-2020-much},  knowledge-graphs (KGs)~\cite{lin-etal-2019-kagnet, feng-etal-2020-scalable, lv2020graph,  saffari-etal-2021-end} or document collections~\cite{chen-etal-2017-reading,izacard-grave-2021-leveraging, guu2020realm, lee-etal-2019-latent, lewis2020retrieval}.
In retrieval-based systems,   differential retrieval can be combined with extractive question answering, as in  REALM~\cite{guu2020realm} and ORQA~\cite{lee-etal-2019-latent},  as well as with generative answer generation, as in RAG~\cite{lewis2020retrieval}.  
%
%


\noindent\textbf{VQA Systems.} Modelling vision and language is central to VQA.
Models can aggregate visual and textual features via cross-modality fusion~\cite{yu2018beyond, singh2019towards,yu2019deep, jiang2020defense, guo2021bilinear}.
Systems can also be pre-trained on large vision-and-language collections~\cite{jia2021scaling} and then fine-tuned for VQA tasks~\cite{tan-bansal-2019-lxmert, chen2020uniter, gan2020large, li2020oscar,  wang2022simvlm, zhang2021vinvl, li-etal-2021-unimo} with VQA datasets such as VQA 2.0~\cite{antol2015vqa}.

\noindent\textbf{Knowledge-based VQA Systems.} KB-VQA can access both structured data, such as ConceptNet and other KGs~\cite{narasimhan2018out,garderes2020conceptbert,li2020boosting,wu2022multi, marino2021krisp}, as well as unstructured data such as Wikipedia passages~\cite{ wu2022multi, gao2022transform, gui2021kat}.    A variety of multimodal approaches have been explored to access external knowledge.  
ConceptBERT~\cite{garderes2020conceptbert} uses attention  to aggregate graph node embeddings from ConceptNet.
KRISP~\cite{marino2021krisp} uses a ``symbolic knowledge module'' to match ConceptNet KG entities with language/visual elements in questions.
MAVEx~\citep{wu2022multi} uses multiple information sources (Google Images, Wikipedia sentences, and ConceptNet) to validate promising answer candidates.
VRR~\citep{luo-etal-2021-weakly} uses Google Search in a retriever-reader pipeline to perform open-ended answer generation.

We also note unpublished contemporaneous work on OK-VQA at the time of submission.
TRiG~\cite{gao2022transform} shows that it is feasible to transform images into textual features for VQA. The features used are similar to those presented here,
although without an emphasis on the role of knowledge retrieval.
PICa~\cite{yang2022empirical} `prompts' GPT-3 with descriptive captions generated from images,
and KAT~\cite{gui2021kat}  exploits an ensemble of DPR, T5, and GPT-3 to 
improve OK-VQA performance.

\section{Methodology}
\label{sec:method}
\begin{figure*}[!h]
    \centering
    \includegraphics[width=1.05\textwidth]{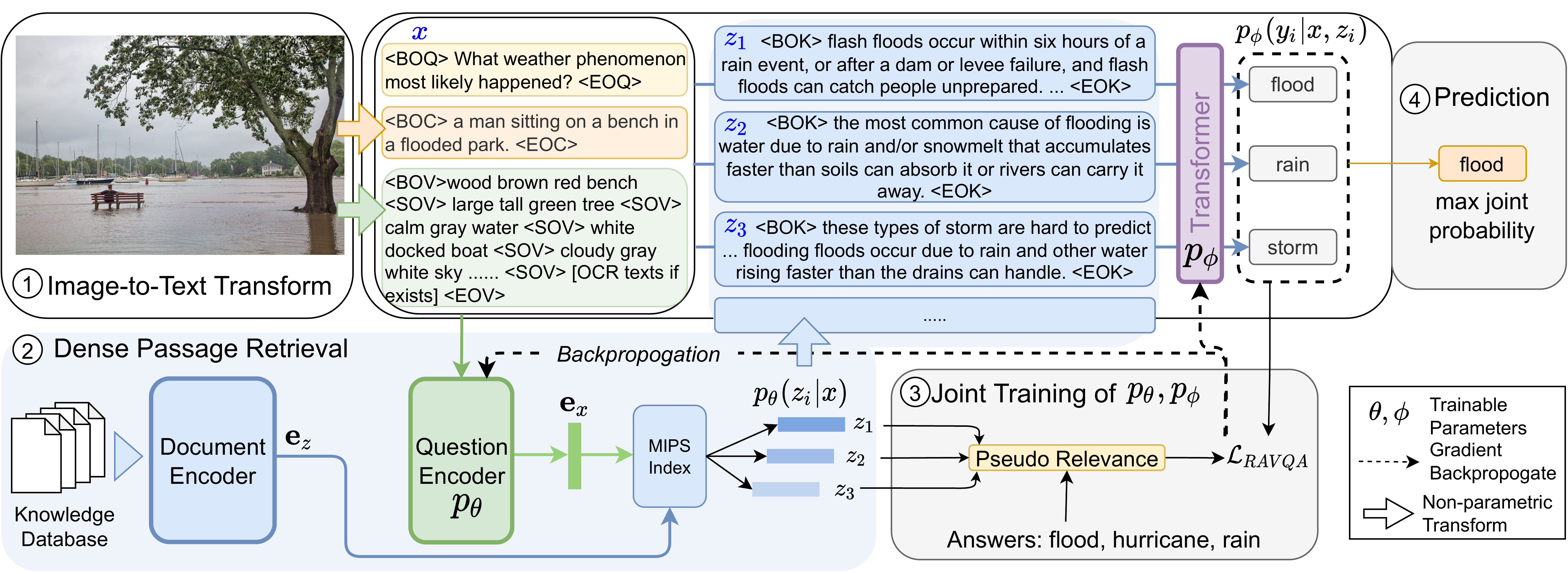}
    \caption{Model overview. (1) Using object detection/image captioning/Optical Character Recognition to transform visual signals into language space.
    (2) Dense Passage Retrieval retrieves documents that are expected to be helpful from the knowledge database; (3) Training the retriever $p_\theta$ and the answer generator $p_\phi$ together using our proposed RA-VQA loss.
    (4) The answer with highest joint probability $p_\theta(z_i|x)p_\phi(y_i|x, z_i)$ is selected.}
    \label{fig:overview}
    \vspace{-0.5cm}
\end{figure*}

We present our  RA-VQA framework that consists of: (1) Vision-to-Language Transformation (Sec.~\ref{sec:method:transform}); (2) Weakly-supervised Dense Passage Retrieval (Sec.~\ref{sec:method:dpr}); (3) Joint Training of Retrieval and Answer Generation (Sec.~\ref{sec:method:joint_training}).

\subsection{Vision-to-Language Transformation}
\label{sec:method:transform}

Prior work has established that images can be transformed into text such that large pre-trained language-based Transformers (e.g. BERT~\cite{devlin-etal-2019-bert}, GPT-2~\cite{radford2019language}, and T5) can be applied to VQA tasks~\cite{luo-etal-2021-weakly, yang2022empirical}.
Systems can be based on straightforward image caption, but we have found improvements by introducing additional visually-grounded features.
In RA-VQA, each image is represented by visual objects and their attributes, image caption, and any text strings detected within the image.
We use an object detection model VinVL~\cite{zhang2021vinvl} that was pre-trained on large object detection datasets to extract visual elements and their attributes (e.g. color and material).

Formally, for an image $I$   we use  VinVL to extract a set of visual objects $\{o_{i}\}$, along with a set of text attributes for each visual object $\{a_{i,j}\}$. 
Visual objects and their attributes are extracted by VinVL at confidence thresholds $0.8$ and $0.6$, respectively. 

Image captioning is performed to extract relationships and interactions among visual elements such as ``a woman holding a knife cuts a cake''.
The pre-trained captioning model Oscar+~\cite{zhang2021vinvl} is applied to process visual features extracted from the VinVL model to generate a caption for the image.
To answer questions related to text strings in images (e.g. ``which language is the book written in?''), Google OCR (Optical Character Recognition) APIs are used to extract text strings from each image.

Hence, a VQA training set $\{(I,q,\mathcal S)\}$, where $\mathcal S$ is a set of answers to a question $q$ about $I$,  can be transformed into a text-only training set ${\mathcal T}=\{(x,  \mathcal S)\}$ that we  use for RA-VQA. 
The string $x$ contains all the text features extracted from the image (the question, the textual attributes for each identified visual object, the generated caption, and any OCR'd text), with special tokens marking the start and end of each type of feature (Fig.~\ref{fig:overview}).   

\subsection{Weakly-supervised Dense Passage Retrieval}
\label{sec:method:dpr}

Dense Passage Retrieval in RA-VQA consists of a query encoder $\mathcal{F}_{q}$ and a document encoder $\mathcal{F}_{d}$, both as Transformer-like encoders.
The aim is to retrieve $K$ documents from an external knowledge database $\ccalZ=\{z_i\}_{i=1}^{N^{d}}$ (e.g. Wikipedia passages) that are expected to be useful for answering a  question.
DPR encodes questions and documents separately into dense feature vectors 
 $   \mathcal{F}_{q}(x) \in \bbR^{h}$ and 
  $   \mathcal{F}_{d}(z) \in \bbR^{h} $.
A scoring function is used to retrieve  documents for each question as the inner product between the representations of $x$ and $z$
\begin{equation}
    {r}(x, z) = \mathcal{F}^\top_{q}(x)   \mathcal{F}_{d}(z)
\end{equation}
RA-VQA training aims to maximize ${r}(x, z)$ when document $z$ is relevant to answering the question.  As discussed in Sec.~\ref{s:intro},
the relevance between $q$ and $z$ cannot be easily obtained and  ``{pseudo relevance labels}''  serve as a proxy.
We use a pseudo relevance function $H(z, \mathcal S)$ which is 1 if $z$ contains an answer in $\mathcal S$ (by string match), and 0 otherwise.

For each question-answer pair $(x,\mathcal S)$ one positive document $z^{+}(x)$ is extracted for training.
In-batch negative sampling is used:
all documents in a training batch other than
{$z^{+}(x)$}
are considered to be negative for $(x,\mathcal S)$~\cite{karpukhin-etal-2020-dense}.
Denoting the negative documents as $\ccalN(x,\mathcal S)$ and the score of the positive document as $\widehat{r}^{+}(x)$ leads to the DPR loss $\ccalL_{DPR} $~:
\begin{equation}
   - \sum_{(x,\mathcal S) \in \mathcal T} \log{\frac{\exp{(\widehat{r}^{+}(x))}}{\exp{(\widehat{r}^{+}(x))}+\hspace*{-2ex}\displaystyle\sum_{z\in \ccalN(x,\mathcal S)}\exp{(\widehat{r}(x,z))}}}
   \label{eq:dprloss}
   \vspace{-0.4cm}
\end{equation}

\subsection{RA-VQA: Joint Training of Document Retrieval and Answer Generation}
\label{sec:method:joint_training}
Given a full query string $x$ extracted from the image-question pair $(I,q)$, DPR returns the $K$  highest scoring documents $\{z_{k}\}_{k=1}^{K}$.
The score assigned by the document retriever {$p_\theta(\cdot|x)$} to a retrieved document is
\begin{equation}
    p_\theta(z_k|x) = \frac{\exp(\widehat{r}(x, z_k))}{\sum_{j=1}^{K} \exp(\widehat{r}(x, z_j))}
\vspace{-0.2cm}
\end{equation}

Open-ended answer generation for each retrieved document $z_k$ is performed with a generative model, such as T5, with parameters $\phi$:
\begin{equation}
    y_k = \argmax_y p_\phi(y|x, z_k)
    \vspace{-0.2cm}
\end{equation}

For each document $z_k$ retrieved for a training item $(x, \mathcal S)$, we train the answer generator to produce the answer string $s^\ast_k$ from the concatenation of $x$ and $z_k$ (as shown in Fig.~\ref{fig:overview}).
We select the most popular\footnote{There are 5 annotators for each OKVQA question. The popularity of an answer is measured by the number of annotators who voted for it.} human response $s^\ast_k$ from $\mathcal S$ such that $s^\ast_k$ is contained in $z_k$;  in the case that $z_k$ does not contain any answer, the most popular answer $s^\ast \in \ccalS$ is selected $s^\ast_k=s^\ast$.
Through this design, we customize the generation target $s^\ast_k$ for each retrieved document instead of training all $(x, z_k)$ pairs towards the most popular human response $s^\ast$. This has been proved to improve the system performance (Appendix~\ref{sec:appendix:full_table}).

We identify two subsets of the retrieved documents $\{z_{k}\}_{k=1}^{K}$ based on pseudo relevance labels and model predictions:
\begin{equation}
\begin{aligned}
\ccalP^{+}(x, \mathcal S) & = \{ k :  y_k = s_k^\ast  \land H(z_k, \mathcal S)=1 \}; \\
\ccalP^{-}(x, \mathcal S) &= \{ k : y_k  \ne s_k^\ast \land  H(z_k, \mathcal S)=0\}.
\end{aligned}
\end{equation}
$\ccalP^{+}$ are indices of pseudo relevant documents that also help the model generate popular answers whereas  $\ccalP^{-}$ identifies documents not expected to benefit answer generation. 
In joint training, we intend to increase the scores of documents in $\ccalP^{+}$ while decreasing the scores for those in $\ccalP^{-}$.
$z_k$ will be put into the negative set if it does not contain any answer ($H(z_k, S) = 0$) and the generation is incorrect ($y_k \ne s_k^*$).\footnote{Note that in this case $H(z_k, S) = 0$ already implies that $z_k$ does not contain any answer and thus $s_k^*=s^*$.} 
This is motivated by our intention to reduce scores for those documents that contain no answers and fail to answer questions.

Formally, joint training of retrieval and answer generation is achieved with a loss  $\ccalL_{RA-VQA}$  that reflects both model predictions and pseudo relevance: 
\vspace{-0.2cm}
\begin{equation}
\begin{aligned}
 -\hspace*{-1ex}&\sum_{(x, \mathcal S) \in \mathcal T} \big( 
 \sum_{k=1}^K \log p_\phi(s^\ast_k|x, z_k) \\ 
 \hspace*{2em}&+ \hspace*{-3ex}\sum_{k \in \ccalP^{+}(x, \mathcal S)}  \hspace*{-3ex} \log p_\theta(z_k|x) 
    - \hspace*{-3ex} \sum_{k \in \ccalP^{-}(x, \mathcal S)} \hspace*{-3ex} \log p_\theta(z_k|x)~\big)
\end{aligned}
\label{eqn:new_rag_loss}
\vspace{-0.25cm}
\end{equation}

The first term in the loss improves answer generation from queries and retrieved documents, taken together.
The remaining terms affect  document retrieval:    
the second term encourages retrieval of documents that are not only pseudo relevant but also lead to production of correct answers, while the third term works to remove irrelevant items from the top ranked retrieved documents.
\begin{figure}[!h]
    \centering
    \includegraphics[width=\linewidth]{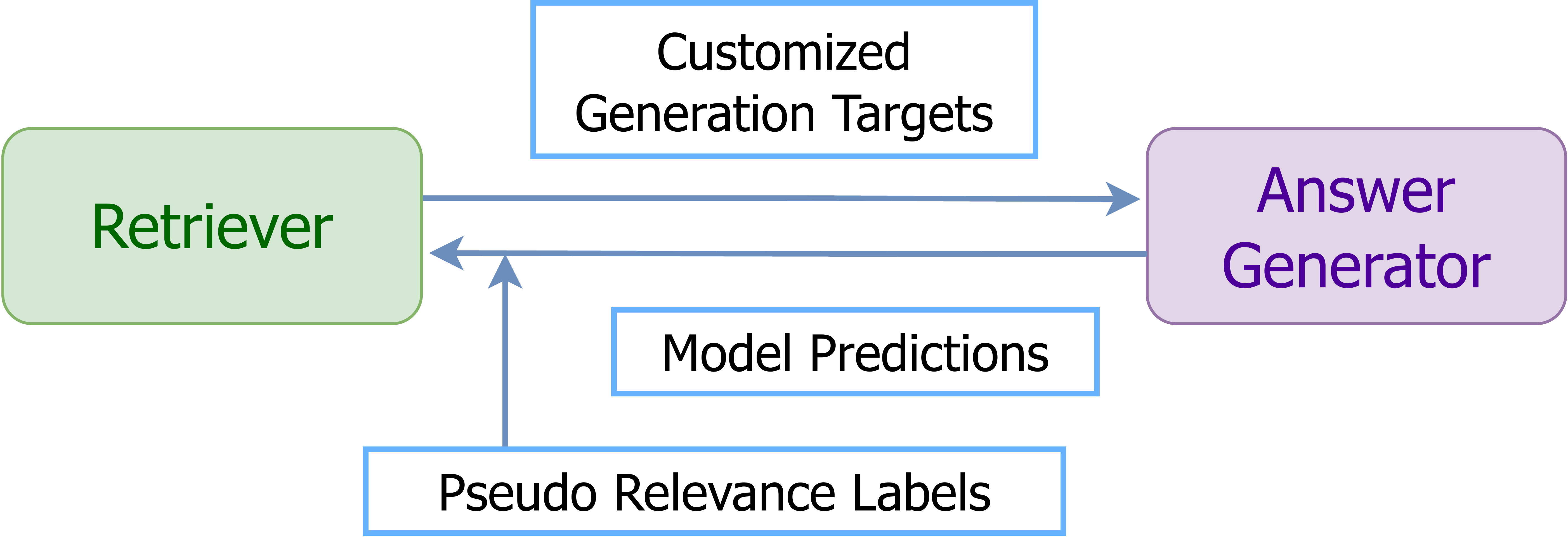}
    \caption{Information flow between the retriever and the answer generator.}
    \label{fig:RAVQA:information_flow}
\end{figure}
The information flow is demonstrated in Fig.~\ref{fig:RAVQA:information_flow}.
Retrieval and generation complement each other in training:  pseudo relevance labels and model predictions provide positive and negative signals to improve retrieval, 
and the improved retrieval leads to improved answer generation by training towards $s^\ast_k$, a customized generation target for each retrieved document $z_k$.

\subsection{RA-VQA Generation}
Given an image query $(I, q)$, a full query $x$ is created (Sec.~\ref{sec:method:transform}) and 
answer generation searches for the answer with the highest joint probability:
\begin{equation}
\begin{aligned}
 & \{z_{k}\}_{k=1}^{K}   = {\argmax_{z}}^K p_\theta(z|x) \\
     \widehat{y},\widehat{z} & = \argmax_{y, z_k} p_\phi(y|x, z_k) ~ p_\theta(z_k|x)
\end{aligned}
\label{eq:ravqagen}
\end{equation}
Answers reflect both generation and retrieval models and retrieval confidence plays a strong role, unlike some prior work such as \citet{luo-etal-2021-weakly}.

\subsection{Pre-Computed FAISS Document Indices}
\label{sec:method:remarks}
Since repeated computation of embeddings for all documents is costly,  we follow ~\citet{lewis2020retrieval} who find that it is enough to train only the question encoder $\mathcal F_q$ and leave document encoder $\mathcal F_d$ fixed.    
%
As shown in Fig.~\ref{fig:overview}, document embeddings are pre-extracted with a pre-trained DPR document encoder.
The FAISS system~\cite{johnson2019billion} is used to index all document embeddings which enables fast nearest neighbour search with sub-linear time complexity.
In training, question embeddings are generated dynamically and documents with highest scores are retrieved using the pre-computed index.

\vspace{-0.1cm}
\section{Experiments}
\vspace{-0.1cm}
\subsection{Datasets and RA-VQA  Configurations}
\label{sec:exp:datasets_and_config}

OK-VQA~\cite{marino2019ok} is currently the largest knowledge-based VQA dataset.
It consists of 14,031 images and 14,055 questions. These questions are split into a training set (9,009 questions) and a test set (5046 questions).
In addition to understanding images and questions,  external knowledge sources are needed to answer questions.

As outside knowledge we use the knowledge corpus collected by~\citet{luo-etal-2021-weakly} from Google Search.
We use the corpus \texttt{GS-full} which consists of 168,306 documents covering training and test questions.
In Appendix~\ref{sec:appendix:full_table} we also report on \texttt{GS-train}, which contains documents relevant to OK-VQA training set questions only. 

\textit{Pre-training:} We start with pre-trained versions of BERT-base and T5-large as the document retriever and the answer generator, respectively.
The retriever was refined by training it on \texttt{GS-full} under the DPR loss (Equation~\ref{eq:dprloss}) with pseudo relevance labels released by~\citet{luo-etal-2021-weakly}.
The already strong retriever serves as a good starting point for all DPR-based models presented in this paper (including RA-VQA and our replication of baselines in the literature).

\textit{OK-VQA Fine-tuning:}  
Our \textbf{RA-VQA} framework trains the answer generator and the retriever jointly under Equation~\ref{eqn:new_rag_loss}. 

We also report on variants of RA-VQA, to investigate the contribution of various model components to overall performance: \\
{\bf RA-VQA-NoDPR} omits retrieval entirely so that answers are generated by the fine-tuned T5 alone.  RA-VQA generation in Equation~\ref{eq:ravqagen} simplifies to
\begin{equation}
    \widehat y_{NoDPR} = \argmax_y p_\phi(y|x)
    \vspace{-0.3cm}
\end{equation}
{\bf RA-VQA-FrDPR}  leaves the retriever frozen after pre-training and fine-tunes the generator only. \\
{\bf RA-VQA-NoPR} is a version of RA-VQA in which document retrieval is trained only with model predictions.
The loss function is as Equation~\ref{eqn:new_rag_loss}, but with positive and negative  document sets defined as
\vspace{-0.2cm}
\begin{equation}
\begin{aligned}
\ccalP_{NoPR}^{+}(x, \mathcal S) & = \{k :  y_k = s_k^\ast \}; \\
\ccalP_{NoPR}^{-}(x, \mathcal S) &= \{k: y_k \ne s_k^\ast \}.
\end{aligned}
\vspace{-0.0cm}
\end{equation}
\textbf{RA-VQA-NoCT} replaces the customized generation targets by the single most popular response ($s_k^\ast$ becomes $s^\ast$ in Equation~\ref{eqn:new_rag_loss}) so that the generator is trained to produce the same answer from every retrieved document.
%

\subsection{Evaluation}
The following metrics are applied to assess the quality of individual answers generated and documents retrieved.
Average scores are then computed over the evaluation set. {The average of 3 runs with different seeds is reported.}
\subsubsection{Answer Evaluation}
\textbf{VQA Score}: We follow~\citet{marino2019ok} to compute VQA Scores using pre-processed human annotations ${\mathcal S}$:
\vspace{-0.2cm}
\begin{equation}
\text{VQAScore}(y,\mathcal S) = \min\big(\frac{\#_{\mathcal S}(y)}{3}, 1\big),
\vspace{-0.2cm}
\end{equation}
{where $\#_{\mathcal S}(y)$ is the number of annotators who answered $y$.} 
This score ensures that a model is  partially rewarded even if it generates one of the less popular answers from amongst the human responses.

\textbf{Exact Match (EM)} 
treats annotated answers equally:
$\text{EM}(y,\mathcal S) = \min( \#_{\mathcal S}(y), 1)$~.


\subsubsection{Retrieval Evaluation}

Following \citet{luo-etal-2021-weakly}, we use pseudo relevance to ascertain whether the retrieved documents are relevant to the response.
It concerns pseudo relevance instead of actual relevance but is still a reasonable metric for retrieval evaluation.


\textbf{Pseudo Relevance Recall (PRRecall)@K} measures how likely the retrieved $K$ documents contains at least one positive document:
\vspace{-0.2cm}
\begin{equation}
\text{PRRecall@K}= \min\big(\sum_{k=1}^{K}H(z_k,\ccalS), 1\big).
\end{equation}


\subsubsection{Integrated System Evaluation}
The above methods evaluate retrieval and answer generation as separate processes.   
We propose additional metrics that assess how the two processes behave in an integrated VQA system.


The \textbf{Hit Success Ratio (HSR)} counts questions that require external knowledge to answer:
\vspace{-0.1cm}
\begin{equation}
 HSR = \mathbbm{1}\big\{ \widehat{y}\in \ccalS \land  \widehat{y}_{NoDPR} \notin \ccalS \big\}.
 \vspace{-0.1cm}
\end{equation}
HSR reflects the value of incorporating external documents into answer generation.

By contrast, \textbf{Free Success Rate (FSR)} counts questions that can be answered without external knowledge.   
\vspace{-0.1cm}
\begin{equation}
 FSR = \mathbbm{1}\big\{ \widehat{y}\in \ccalS \land  \widehat{y}_{NoDPR} \in \ccalS \big\}.
 \vspace{-0.1cm}
\end{equation}
A high FSR suggests a model can generate correct answers `freely' without being distracted by retrieved documents if they are not needed.

We also  assess performance as a function of the number of documents retrieved during training and testing, 
$\mathbf{K_{\text{\textbf{train}}}}$ {and} $\mathbf{K_{\text{\textbf{test}}}}$.
In practice, $K_{\text{train}}$ has the greater effect on GPU usage, since a large $K_{\text{train}}$ requires at least $K_{\text{train}}$ forward passes for each question and an Adam-like optimizer must compute and store the associated gradients~\cite{KingmaB14adam}.
In contrast, GPU memory required during testing is significantly less, as there is no optimizer involved.
We are in particular interested in the ability of knowledge-augmented systems that can be robustly trained with small $K_{\text{train}}$ while yielding improved performance with  large $K_{\text{test}}$.



\subsection{Baseline Systems}

\subsubsection{Retrieval Augmented Generation}
RAG~\cite{lewis2020retrieval} 
is based on DPR and an answer generator that are trained jointly  
by approximately marginalizing the probability of $y$ over the retrieved documents.
In the notation of Sec.~\ref{sec:method}:
\vspace{-0.1cm}
\begin{equation}
\begin{aligned}
    p_{RAG}(y|x) 
    \approx& \sum_{k=1}^{K} p_\phi(y|x, z_k) ~p_\theta(z_k|x)
\end{aligned}
\vspace{-0.3cm}
\end{equation}
{The answer generator and the retriever are jointly trained by optimizing the RAG loss: $-\sum_{(x, \mathcal S) \in \mathcal T} \log\big(p_{RAG}(s^\ast|x)\big)$.
}
The rationale is that $p_{\theta}(z_k|x)$ will increase if $z_k$ has a positive impact on answer generation~\cite{lewis2020retrieval}.
We consider RAG as an important baseline and have carefully replicated its published implementation\footnote{\href{https://github.com/huggingface/transformers/tree/master/examples/research_projects/rag}{The authors released RAG in huggingface.}}.

\subsubsection{Baseline Systems in the Literature} 
\begin{table*}[!h]
\centering
\small
\begin{tabular}{lcclllccccc}
\toprule
Model                   & T5 & GPT-3           & $K_{\text{train}}$ & $K_{\text{test}}$ & Knowl. Src.       & PRRecall  & HSR~/~FSR & H/F & EM                                & VQA\\
\midrule
 ConceptBERT             &     $\times$ & $\times$      &  -   & -    & C                                         &&&&      & 33.66      \\
 KRISP                   &     $\times$ & $\times$                       &    -      &  -       & C + W &&&& & 38.35      \\
 VRR & $\times$ & $\times$                        & 100      & 100     & GS                   &&&&                          & 45.08      \\
 MAVEx                   &     $\times$ & $\times$    & -   &     -    & W + C + GI                  &&&& & 39.40       \\
 KAT-T5                     & \checkmark & $\times$              & 40       & 40      & W                        &&&&                         & 44.25      \\
TRiG    & \checkmark & $\times$              & 5      & 5     & W                                                &&&& 49.21 & 45.51       \\
TRiG    & \checkmark & $\times$              & 100      & 100     & W                                            &&&&  53.59  & 49.35       \\
TRiG-Ensemble    & \checkmark & $\times$        & 100      & 100     & W                                        &&&&    54.73    & 50.50       \\
TRiG*                   & \checkmark & $\times$              & 5        & 5       & GS                            &  &   & &  52.79             & 48.32      \\
RAG*  &   \checkmark & $\times$  & 5 & 5 & GS & 82.34 & 12.28~/~40.24 & 0.31 &  52.52 & 48.22\\
RA-VQA (Ours)                     & \checkmark & $\times$              & 5        & 5       & GS                    & 82.84 & 16.75~/~41.97 & 0.40 &  58.72    & 53.81  \\
RA-VQA (Ours)                     & \checkmark & $\times$              & 5        & 50     & GS    & 96.55                   & 17.32~/~42.09 &  \textbf{0.41} &               \textbf{59.41}  & \textbf{54.48}      \\
\midrule
\textit{Ablation Study}\\
\midrule
RA-VQA-FrDPR                    & \checkmark & $\times$              & 5        & 5       & GS  & 81.25 & 15.01~/~40.76 & 0.37 & 55.77   & 51.22      \\
RA-VQA-NoPR                    & \checkmark & $\times$              & 5        & 5       & GS  & 77.67 & 15.97~/~41.83 & 0.38
 &  57.80  & 52.98      \\
RA-VQA-NoCT                    & \checkmark & $\times$              & 5        & 5       & GS  & 83.77 & 14.55~/~42.96 & 0.33
 &  57.51  & 52.67      \\

\midrule
\multicolumn{7}{l}{\textit{GPT-3-based Systems ($>$175 Billion Parameters)}}\\
\midrule
 PICa                    & $\times$ & \checkmark                &     -     & -        & GPT-3                 &&&&                              & 48.00         \\
KAT-Knowledge-T5            & \checkmark & \checkmark &    40      &   40      & W + GPT-3                                             &&&&    & 51.97      \\
KAT-Ensemble                     & \checkmark & \checkmark & 40       & 40      & W + GPT-3                                                &&&& & 54.41      \\
\bottomrule
\end{tabular}
\caption{RA-VQA vs. Baseline Systems.
Knowledge Sources: {\bf\underline C}onceptNet; {\bf\underline W}ikipedia; {\bf\underline G}oogle {\bf\underline S}earch; {\bf\underline G}oogle {\bf\underline I}mages; {\bf\underline{GPT-3}} closed book knowledge.
\underline{H/F}: HSR to FSR ratio.
PRRecall, HSR, FSR, and EM are reported in percentage (\%).
PRRecall is reported at the corresponding $K_{\text{test}}$.
}
\label{tab:comparison_with_baselines}
\vspace{-0.3cm}
\end{table*}

We compare against the published OK-VQA results from systems described in Sec.~\ref{sec:related_work}: \textbf{ConceptBERT}, \textbf{KRISP}, \textbf{MAVEx}, and \textbf{VRR}. 
We also report performance against unpublished (non peer-reviewed) systems \textbf{TRiG}\footnote{At the time of submission, TRiG has not been published.}, \textbf{PICa}, and \textbf{KAT}.
\textbf{TRiG} uses a similar image-to-text transform as this work, so to enable fair comparison with our model we replicate their knowledge fusing method with our features.
Baseline results are reported in~Table~\ref{tab:comparison_with_baselines}; 
baseline results marked  {\bf *} are our own.
{\textbf{TRiG*}, our own implementation of TRiG, concatenates $K$ encoder outputs for the decoder to use in generation.}

We make some  particular observations.  Our {TRiG*}  improves over {the results released in its paper (VQA Score of 48.32 vs 45.51)} at $K_{\text{train}} = K_{\text{test}}=5$;   {TRiG} and {TRiG} Ensemble both benefit from more retrieved documents in training and testing ($K_{\text{train}} = K_{\text{test}}=100$), although at great computational cost.  Best performance with KAT-T5 and VRR similarly requires large document collections in training and in test.   

We include results from GPT-3 based systems because they are amongst the best in the literature, but we note that GPT-3 is so much bigger than T5 (175 billion parameters in GPT-3 v.s. 770 million in T5-large) that simply switching a system implementation from T5 to GPT-3 can give significant improvements:   KAT-T5 achieved a 44.25 VQA Score while {ensembling it with GPT-3} yields 54.41;
and GPT-3 alone already achieved good performance with prompting (PICa with 48.00 VQA Score).     Our RA-VQA system is based on T5, but we still find competitive results even in comparison to systems incorporating GPT-3 (54.48 vs 54.41 of KAT-Ensemble). 
\subsection{RA-VQA Performance  Analysis}

We find that RA-VQA matches or improves over all  baseline systems with a VQA Score of 54.48.
This is with a configuration of $K_{\text{train}}=5$ and $K_{\text{test}}=50$,  
thus validating our claim that RA-VQA can use a large number of retrieved documents in testing (50) while using relatively few retrieved documents in training (5).
We find that reducing the number of retrieved documents in test ($K_{\text{test}}=5$) reduces the VQA Score,  but still yields performance better than all baselines except the KAT ensemble.    

We also find that RA-VQA performs well relative to GPT-3 baselines.  
RA-VQA yields a higher VQA score than  KAT-Knowledge-T5  (54.48 vs. 51.97) and matches the KAT-Ensemble system.
We emphasize that  RA-VQA is significantly smaller in terms of parameters (and in model pre-training data) than these GPT-3 based systems and that training RA-VQA requires much less memory ($K_{\text{train}}=5$ vs $K_{\text{train}}=40$).

%
%

%

\subsubsection{Contributions of Query Features and DPR to Overall Performance}
\begin{table}[!h]
\small
\centering
\begin{tabular}{lcccccc}
\toprule
Model   & Q  & O & A & C & T            & VQA Score \\
\midrule
RA-VQA-NoDPR & \checkmark & $\times$&$\times$&$\times$&      $\times$              & 28.05     \\
RA-VQA-NoDPR & \checkmark & \checkmark &$\times$&$\times$&    $\times$             & 40.95     \\
RA-VQA-NoDPR & \checkmark & \checkmark &\checkmark&$\times$&    $\times$            & 42.14     \\
RA-VQA-NoDPR & \checkmark & \checkmark &\checkmark& \checkmark&     $\times$         & 45.31     \\
RA-VQA-NoDPR & \checkmark & \checkmark &\checkmark& \checkmark& \checkmark          & 46.16     \\ 
RA-VQA-FrDPR & \checkmark & \checkmark &\checkmark& \checkmark& \checkmark  & 51.22\\
RA-VQA  & \checkmark & \checkmark &\checkmark& \checkmark& \checkmark   & 53.81    \\
\bottomrule
\end{tabular}
\caption{Ablation study on input features and system configurations: {\bf \underline Q}uestions; {\bf \underline O}bjects; {\bf \underline A}ttributes associated with objects; {\bf \underline C}aptions;  visible {\bf \underline T}ext from OCR.  $K=5$ in RA-VQA and RA-VQA-FrDPR.}  
\label{tab:input_feature_ablation}
\vspace{-0.4cm}
\end{table}
A detailed ablation study on input features is presented in Table~\ref{tab:input_feature_ablation}.
As shown, the T5 model fine-tuned on OK-VQA achieves a 28.05 VQA Score.
The VQA Score increases to 46.16 as objects, object attributes, image captions, and OCR texts are incorporated into RA-VQA-NoDPR.
With $5$ retrieved documents, RA-VQA-FrDPR yields a 51.22 VQA Score, with a further improvement (53.81 VQA Score) in full training of retrieval and answer generation, confirming that outside knowledge is needed to answer OK-VQA questions.

\subsubsection{Benefits of Integrated Training}
\label{sec:experiments:rag}

Joint training is a key benefit of our proposed RA-VQA framework:  model predictions combine with pseudo relevance labels to improve retrieval, and the resulting improved retrieval in turn provides customized answer generation targets.
To quantify these effects,
we take RA-VQA-FrDPR as a starting point (Table~\ref{tab:comparison_with_baselines}).  Comparing it with other RA-VQA models suggests that DPR training in itself is necessary, as using only pre-trained DPR (RA-VQA-FrDPR) leads to weaker VQA Score (51.22).
Using model predictions alone in joint DPR training (RA-VQA-NoPR) leads to a higher VQA Score (52.98 vs 51.22), but a significantly lower PRRecall (77.67\% vs 81.25\%). 
The model decides to remove some pseudo relevant documents but achieves better performance.
This points to a potential problem that can arise.  Pseudo relevance is only an imperfect indication of true relevance and so is not an ideal criteria on its own.  Training DPR to retrieve pseudo relevant documents could  result in misleading documents being used in answer generation.

Using both pseudo relevance labels and model predictions in DPR training (RA-VQA) improves VQA Score to 53.81 and notably improves PRRecall to 82.84\%.
Including pseudo relevance ensures that potentially useful documents are retained, even while the generator is still learning to use them.

We also note that when generation targets are not customized for each retrieved document (RA-VQA-NoCT), VQA Score drops by 1.14 relative to RA-VQA, showing that customized generation targets play an important role in the overall system:
by training the model to extract the reliable answers available in retrieved documents, answer generation and retrieval are both improved.

\subsubsection{Interaction of Retrieval and Generation}

Table~\ref{tab:comparison_with_baselines} also reports our investigation into the interaction between document retrieval and answer generation.
In comparing RA-VQA-FrDPR (frozen DPR) to RA-VQA, 
we see that joint training of DPR yields not only improved EM but also significantly higher HSR (+1.74\%) and FSR (+1.21\%).
Manual inspection of OK-VQA reveals that there are many general knowledge questions.   For example, document retrieval is not needed to answer  the question "Is this television working?" in reference to a picture of a broken television lying in a field.    A high FSR indicates good performance on such questions.
By contrast,   a high HSR reflects the ability to use document retrieval to answer the questions that truly require external documents.


Both EM and HSR are further improved for $K_\text{test} = 50$ in RA-VQA, with little change in FSR.    The increased HSR to FSR ratio (0.41 vs. 0.40) indicates that RA-VQA is using these additional retrieved documents to answer the questions that need outside knowledge.   
%

HSR and FSR also explain the relatively weak performance of RAG*.
We see that although RAG* and RA-VQA-FrDPR  have similar FSRs, RAG* has higher PRRecall but lower HSR (by -2.73\%).
This suggests  RAG*'s DPR model is not well matched to its answer generator.  The result is that retrieved documents remain unexploited.
In manual examination of gradients of document scores in training, we find anecdotally that adjustments to document scores are often counter-intuitive:
documents that do not contain answers can still have their scores upvoted if 
the answer generator happens to find a correct answer by relying only on the ability of T5 model.
This works against a model's ability to find answers in retrieved documents even when those documents are relevant.

\subsubsection{Effects of $K_{\text{train}}$}
\label{sec:RAVQA:experiments:effects_of_K}
\begin{figure}[!h]
\vspace{-0.4cm}
    \centering
    \includegraphics[width=0.86\linewidth]{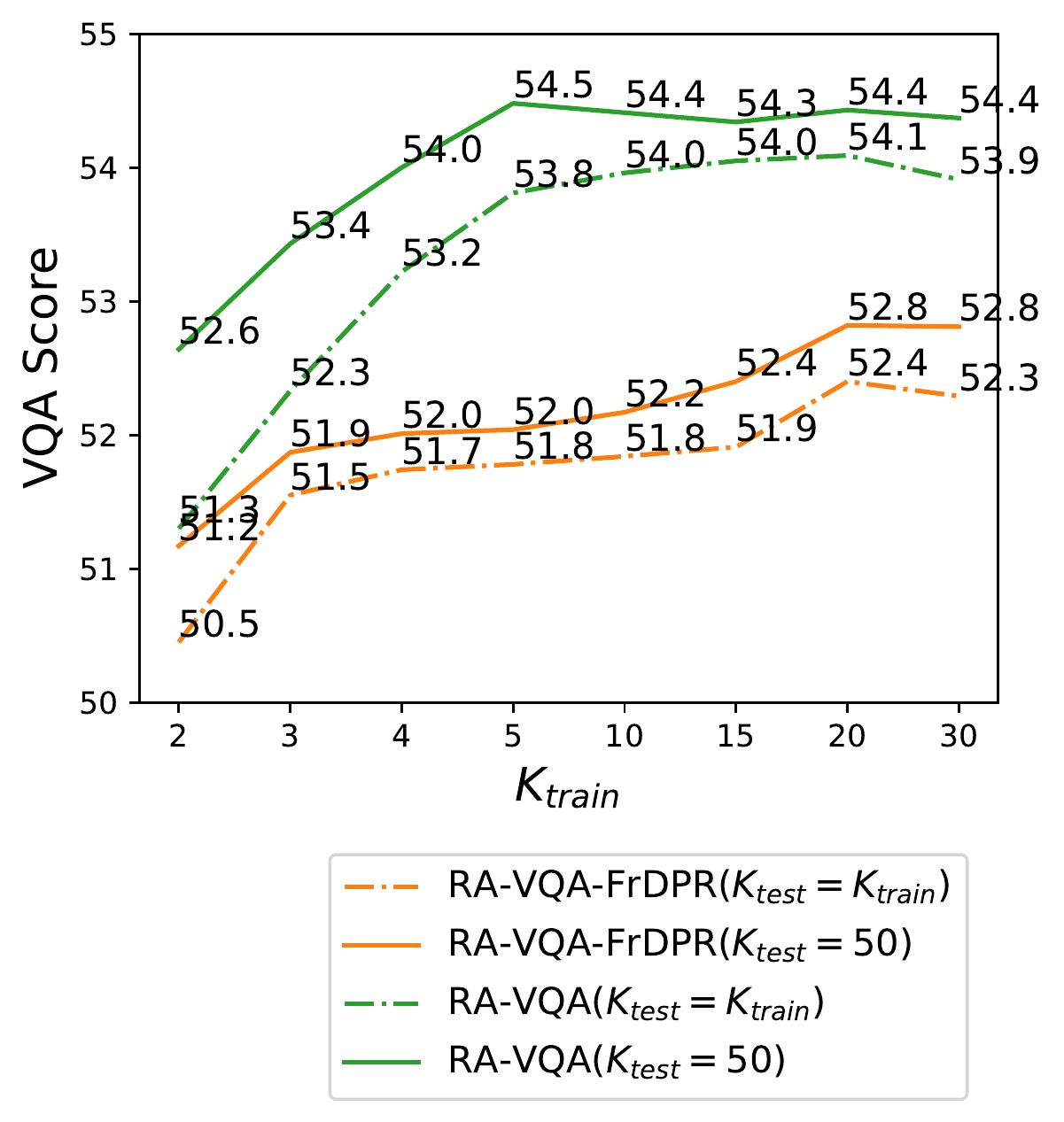}
    \caption{VQA Scores against $K_{\text{train}}$.
    Dashed line: $K_{\text{test}}=K_{\text{train}}$;
    solid line: $K_{\text{test}}=50$.
    Our proposed model achieves the best performance when additional documents are retrieved in test ($K_{\text{test}}=50$). This holds even for models trained to retrieve fewer documents.
    }
    \label{fig:effects_of_k_train}
    \vspace{-0.3cm}
\end{figure}

As noted, retrieving a large collection of documents in training is costly (large $K_\text{train}$).
Fig.~\ref{fig:effects_of_k_train} shows that RA-VQA can be trained with relatively few retrieved documents ($K_\text{train}=5$).
We gradually increase $K_{\text{train}}$ while fixing $K_{\text{test}}=K_{\text{train}}$ (dash lines) and $K_{\text{test}}=50$ (solid lines).
RA-VQA achieves consistent performance ($\sim$54.4 VQA Score) at $K_{\text{train}}\geq5$ and $K_{\text{test}}=50$, which suggests that our joint training scheme is able to gather most useful knowledge into a top-$50$ list even when the model is trained to retrieve fewer documents.
This is not the case for the frozen DPR systems which require increasing $K_{\text{train}}$ to obtain best performance.
RA-VQA's superior performance shows that joint training of retrieval and generation yields clear benefits in computation and answer quality.

We also note that $K_{\text{train}}=5$ is an optimal design choice that strikes a balance between training computational cost and system performance.
The green curves at $K_{\text{train}}<5$ also suggest that it is beneficial to include at least 5 documents in training.
This is because $K_{\text{train}}\geq 5$ offers a good PRRecall (over 80\%), which provides documents of higher quality for training the system.

\subsection{Additional Analyses}
We redirect readers to other interesting analyses (e.g. an analysis of computational cost) and case studies in Appendices~\ref{sec:appendix:supplementary}-\ref{sec:appendix:case_study}.

In addition, in Appendix~\ref{sec:RAVQA:appendix:FVQA_results}, we evaluate our proposed framework on another popular Knowledge-based VQA dataset, FVQA (Fact-based VQA)~\cite{wang2017fvqa}.
Similarly, the RA-VQA framework with joint training achieves better results over the baseline systems with a frozen DPR component, showing the generalizability of our proposed framework.

\section{Conclusion}
Retrieval-Augmented Visual Question Answering is a novel modelling framework for integrated training of DPR and answer generation.   
We have evaluated RA-VQA on the OK-VQA task and we find significantly better performance than the independent training of component system.
Through diagnostic metrics such as HSR and FSR we analysed the interaction between retrieval and generation, and have also shown how RA-VQA's gains arise relative to 
other approaches,  such as RAG.
As a further practical benefit, we found that RA-VQA can be used with larger numbers of retrieved documents than were used in system training, yielding computational savings without sacrificing performance.

The code for this paper will be released at \href{https://github.com/LinWeizheDragon/Retrieval-Augmented-Visual-Question-Answering}{https://github.com/LinWeizheDragon/Retrieval-Augmented-Visual-Question-Answering}.

\section{Acknowledgement}
W. Lin was supported by a Research Studentship funded by Toyota Motor Europe (RG92562(24020)). We thank our colleagues, Daniel Olmeda Reino (Toyota Motor Europe) and Jonas Ambeck (Toyota Motor Europe), who provided insight and expertise in this project.

We thank Zhilin Wang (University of Washington) and Alexandru Coca (University of Cambridge) for comments that greatly improved the manuscript.
We would also like to thank all the reviewers for their knowledgeable reviews.

\section{Limitations}
One possible limitation is that some relevant information (such as relative positioning of objects in the image) could be lost in transforming images independently of the information being sought.  Extracting visual features based on queries could be a natural next step, although query-specific processing of the image collection would be computationally expensive.

We selected the Google Search corpus ~\cite{luo-etal-2021-weakly} as the knowledge base for our question answering system.  Its advantages are that it is large, openly available, and can be readily used to replicate the results in this paper.
However some visual question types (e.g. ‘Is the athlete right or left handed?’)  could plausibly require both complex reasoning and more closely relevant documents from additional knowledge sources (such as Wikipedia).
Our system may be limited with respect to these considerations.

\bibliography{reference}
\bibliographystyle{acl_natbib}

\appendix

\begin{table*}[!htp]
\small
\centering
\begin{tabular}{lcccccccccc}
\toprule
\multirow{2}{*}{Models} & \multicolumn{2}{c}{$K=5$} & \multicolumn{2}{c}{$K=10$} & \multicolumn{2}{c}{$K=20$} & \multicolumn{2}{c}{$K=50$} & \multicolumn{2}{c}{$K_{\text{test}}=5$}                   \\\cmidrule(r){2-3} \cmidrule(r){4-5} \cmidrule(r){6-7} \cmidrule(r){8-9} \cmidrule(r){10-11}
                        & P         & R         & P          & R         & P          & R         & P          & R       & EM      & VQA Score \\
\midrule
VRR~\cite{luo-etal-2021-weakly}       &       -    & 80.40   &    -        & 88.55   &       -     & 93.22   &      -      & \textbf{97.11}   &      -      & 42.54     \\
RA-VQA-FrDPR       & 51.82   & 81.25   & 49.20    & 88.51   & 45.98    & 92.98   & 41.24    & 96.75   &  55.77 & 51.22   \\
RA-VQA                    & \textbf{57.39}   & \textbf{82.84}   & \textbf{54.83}    & \textbf{89.00}   & \textbf{51.48}    & \textbf{93.62}   & \textbf{46.36}    & 96.47  & \textbf{58.72}    &  \textbf{53.81}    \\
\bottomrule
\end{tabular}
\caption{Comparing retrieval performance of VRR and our RA-VQA models. The same knowledge corpus (\texttt{GS-full}) was used. \underline{P}: Pseudo Relevance Precision; \underline{R}: Pseudo Relevance Recall; \underline{EM}: Exact Match. \underline{P} under $K=5$ refers to PRPrec@5. VRR was trained on $K_\text{train}=100$, while RA-VQAs were trained on $K_\text{train}=5$.}
\label{tab:evaluation_retrieval_supplementary}
\vspace{-0.3cm}
\end{table*}

\section{Training Details and Artifacts}
\label{sec:appendix:training}
Adam~\cite{KingmaB14adam} was used in this paper.
In DPR pre-training, the retriever was trained for 6 epochs with a constant learning rate $10^{-5}$.
In RA-VQA training (including RAG*), the learning rates are $10^{-5}$ for the retriever, and $6 \times 10^{-5}$ for the answer generator, linearly decaying to $0$ after $10$ epochs.
In the training of RA-VQA-NoDPR and TRiG*, the initial learning rate is $6\times 10^{-5}$.
Empirically, the checkpoints at epoch 6 were used in testing.
All experiments were run on Nvidia A-100 GPU clusters.
With $K_{\text{train}}=5$, the RA-VQA training takes around 5 hours (10 epochs) while testing takes 5 minutes.
The time cost increases as $K_{\text{train}}$ increases, approximately linearly.

Pre-trained model parameters (e.g. T5-large and BERT-base) are provided by huggingface~\cite{wolf-etal-2020-transformers} accompanied by Python libraries (under Apache License 2.0).
FAISS~\cite{johnson2019billion} is under MIT License.


\section{Supplementary Tables and Figures}
\label{sec:appendix:supplementary}
Limited by space available, we present supplementary tables and figures in this section, offering more findings to readers.

\subsection{Full Version of Table~\ref{tab:comparison_with_baselines}}
\label{sec:appendix:full_table}

We provide the full version of Table~\ref{tab:comparison_with_baselines} in Table~\ref{tab:comparison_with_baselines_full}.
Some more discussions about customized generation target in joint training is provided.

As noted, RA-VQA improves retrieval with the feedback of model predictions, and in turn the improved retrieval leads to improved answer generation by training towards $s_k^\ast$, a customized generation target for each retrieved document $z_k$.
We remove this interaction from RA-VQA models by enforcing $s_k^\ast = s^\ast$ (the most popular human response), independent of the retrieved $z_k$.
The ablated models are denoted with a \texttt{*-NoCT} suffix.

As shown in Table~\ref{tab:comparison_with_baselines_full}, customizing generation targets for each retrieved $z_k$ in training yields performance boost for both RA-VQA-FrDPR and RA-VQA, showing that this supervision signal is beneficial to overall system performance.
We also notice that the improvement to RA-VQA (+1.14 VQA Score) is larger compared to RA-VQA-FrDPR (+0.56 VQA Score), showing that customizing the generation target brings more benefits when the retrieval is improved within our proposed RA-VQA joint training framework.
This further confirms that the two components, retrieval and answer generation, complement each other bi-directionally.

\subsection{Retrieval Performance of RA-VQA}

In addition to Pseudo Relevance Recall (PRRecall) introduced in the paper, we further evaluate retrieval performance with \textbf{Pseudo Relevance Precision (PRPrec)@K}, which is calculated as the rate of pseudo positive documents  in all the $K$ documents retrieved for a question:
\begin{equation}
\text{PRPrec@K}= \frac{1}{K}\sum_{k=1}^{K}H(z_k, \ccalS)
\end{equation}
where $H(\cdot)$ is the pseudo relevance function introduced in Sec.~\ref{sec:method:dpr}.

\begin{figure}[!h]
\centering     
\subfigure[VQA Score vs $K_{\text{test}}$]{\label{fig:VQA_with_K}\includegraphics[width=0.8\linewidth]{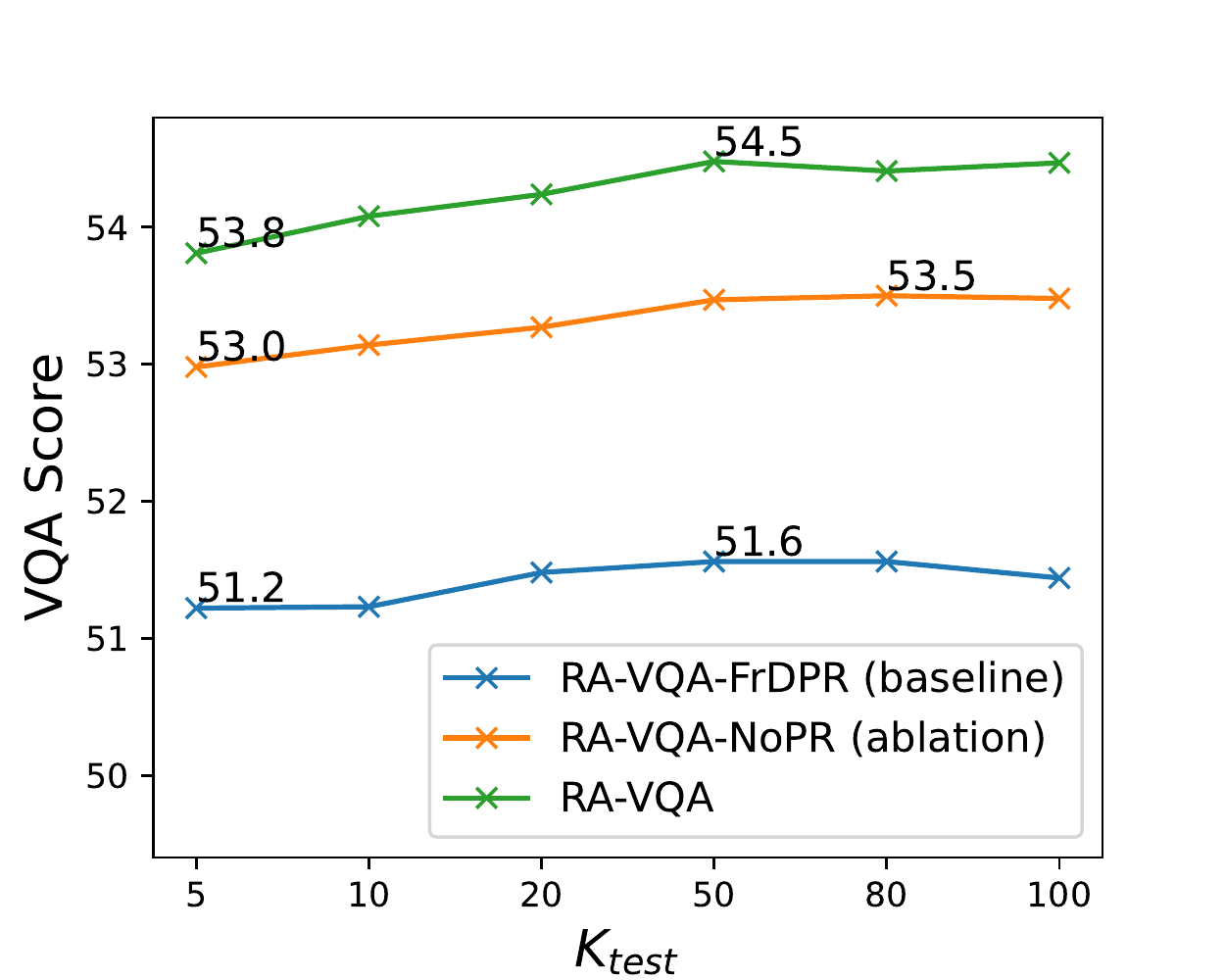}}\\
\subfigure[Recall vs $K_{\text{test}}$ ]{\label{fig:recall_with_K}\includegraphics[width=0.77\linewidth]{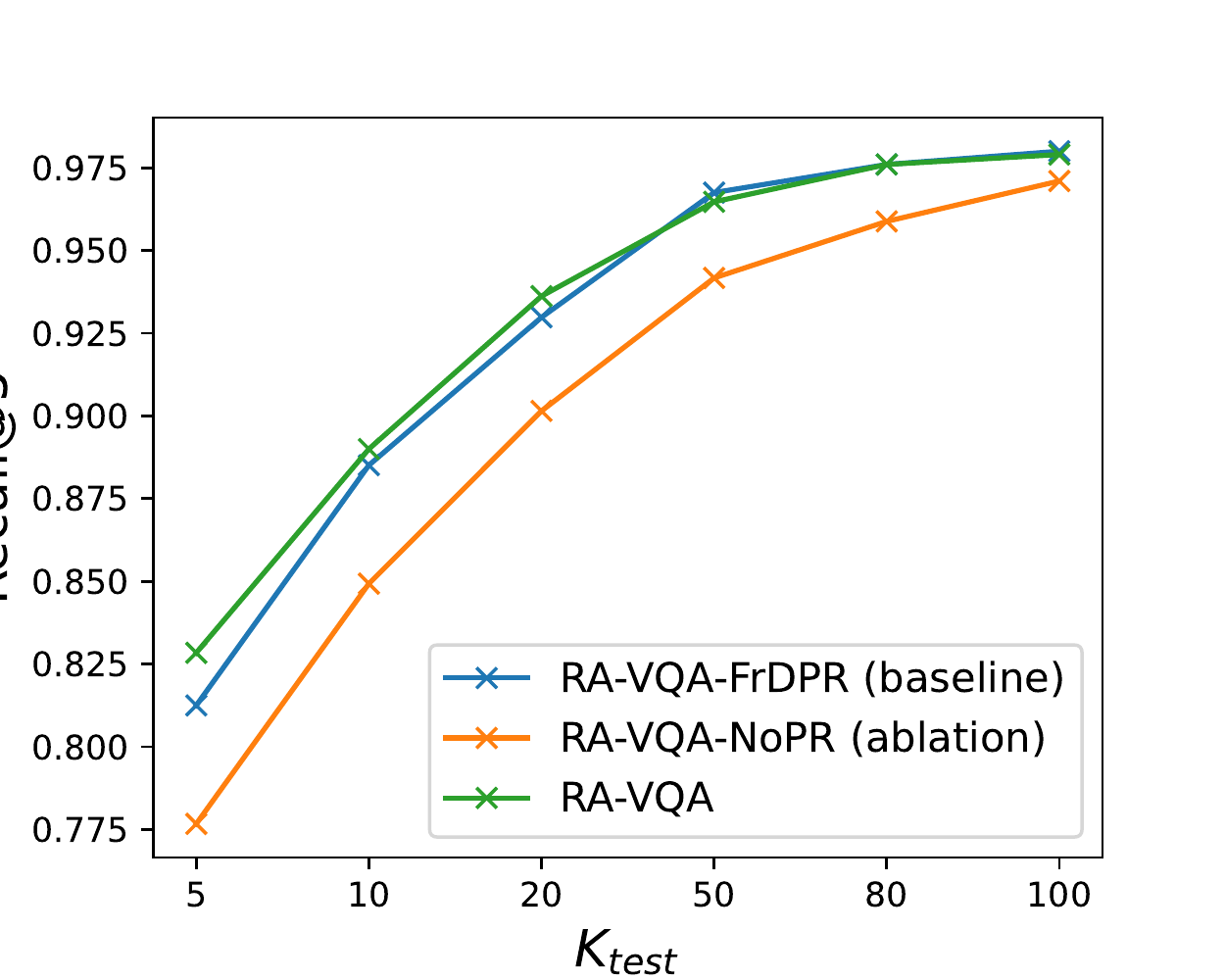}}
\caption{Comparison of model performance as more documents are retrieved in testing. These models are all trained with $K_{\text{train}}=5$.
In RA-VQA full joint training (green), combining model predictions with pseudo relevance labels yields higher PRRecall at low $K_{\text{test}}$, showing that full joint training improves retrieval;
RA-VQA-NoPR (orange), which uses only model predictions in training, achieves a higher VQA Score with lower Pseudo Relevance Recall compared to the RA-VQA-FrDPR with frozen DPR in training (blue), which suggests that Pseudo Relevance is only an approximate measurement of actual relevance.}
\label{fig:generalization_with_K}
\end{figure}

The success of our RA-VQA model can be further explained by Table~\ref{tab:evaluation_retrieval_supplementary}.
As expected, RA-VQA-FrDPR (pre-trained DPR) achieves similar retrieval performance as VRR~\cite{luo-etal-2021-weakly} since they are both based on DPR and are trained with the same pseudo-relevance-based labels.
Our proposed RA-VQA, with a substantial improvement in Recall over RA-VQA-FrDPR (82.84 PRRecall@5 vs 81.25 PRRecall@5), achieves significantly higher Precision (57.39 PRPrec@5 vs 51.82 PRPrec@5).
This also yields substantial improvements to both EM (+3.05\%) and VQA Score (+2.59\%).
This suggests that training the retriever jointly presents more potentially relevant documents to answer generation, improving the quality of the top-ranked documents.

\subsection{Effects of Retrieving More Documents in Test}

Fig.~\ref{fig:generalization_with_K} presents the change of VQA Score and PRRecall as additional documents are retrieved in test (increasing $K_{\text{test}}$).

PRRecall is improved dramatically as $K_{\text{test}}$ increases from 5 to 50, after which only marginal improvement is observed.
Similarly, the VQA Score of these models is improved as more documents are presented in test, and the performance peaks at $K_{\text{test}}\sim50$.
This suggests that including more additional documents in test is more likely to include the truly relevant document to help answer the question yet along with more distracting and misleading documents.

RA-VQA-NoPR (introduced in Sec.~\ref{sec:exp:datasets_and_config}), which uses only model predictions in training to adjust document scores without pseudo relevance labels, yields a significantly lower PRRecall curve (orange curve in Fig.~\ref{fig:recall_with_K}) than RA-VQA-FrDPR (blue curve) across all $K_{\text{test}}$s, but achieves much higher VQA performance (Fig.~\ref{fig:VQA_with_K}).
This further confirms that Pseudo Relevance Labels are a weak signal and a high PRRecall does not necessarily guarantee to gather truly relevant knowledge in retrieval.

\section{Evaluation of Computational Cost}
\begin{table}[!h]
\small
\centering
\begin{tabular}{llll}
\toprule
Model       & GPU memory & Wall time & Batch Size                 \\
\midrule
RAVQA       & 70G             & 180 min                  & 2 $\times$ 16 \\
RAVQA-FrDPR & 67G             & 160 min                  & 4 $\times$ 8  \\
TRiG*       & 70G             & 220 min                  & 16 $\times$ 2    \\
\bottomrule
\end{tabular}
\caption{Comparing computational cost of models. GPU memory: total GPU memory occupied in runtime (approximately); Wall time: the required training time for 8 epochs; Batch Size: \texttt{per\_device\_batch\_size}$\times$ \texttt{gradient\_accumulation\_steps}. The statistics was collected using one Nvidia A100 GPU (80G).}
\label{tab:RAVQA:computational_cost}
\end{table}

As shown in Fig.~\ref{tab:RAVQA:computational_cost}, comparing with the DPR baseline, the wall time is increased by only 20 mins for RAVQA joint training. Our method does not significantly increase the computation needs. In comparing with TRiG, the total training time is also reduced; this is because TRiG concatenates all $K$ hidden states for decoding~\cite{gao2022transform}, which is computationally expensive.

\begin{table}[!h]
\centering
\small
\begin{tabular}{lr}
\toprule
Model      & Wall Time (8 epochs) \\
\midrule
RAVQA $K_{\text{train}}=5$  & 3 hrs     \\
RAVQA $K_{\text{train}}=10$ & 7 hrs     \\
RAVQA $K_{\text{train}}=15$ & 12 hrs   \\
\bottomrule
\end{tabular}
\caption{Wall time of the RAVQA model with an increasing $K_{\text{train}}$. More time is required for training 8 epochs as $K_{\text{train}}$ increases.}
\label{tab:RAVQA:compare_walltime_with_K}
\end{table}

As $K$ increases, the DPR-based system takes significantly more time to train. This is a real issue for other DPR-based systems that use very high $K_{\text{train}}$ (e.g.  TRiG, KAT), but not the case for our framework:  we verified in Sec.\ref{sec:RAVQA:experiments:effects_of_K} that $K=5$ achieves almost the same results as $K\geq 10$, which is a desirable feature that can significantly reduce the required computation cost for achieving the best performance.

\section{Random Guess with OK-VQA Questions}
\label{sec:appendix:random_guess}
\begin{table}[!h]
\centering
\small
\begin{tabular}{lr}
\toprule
Question                                         & Prediction  \\
\midrule
What type of bird is this?                       & hawk        \\
What time of day is it?                          & afternoon   \\
What kind of dog is this?                        & chihuaha    \\
What kind of bird is this?                       & hawk        \\
What sport is this?                              & horse race  \\
What breed of horse is that?                     & clydesdale  \\
What century is this?                            & 19th        \\
What kind of birds are these?                    & pigeon      \\
What city is this?                               & new york    \\
What is the weather like?                        & rainy       \\
What do these animals eat?                       & grass       \\
What activity is this?                           & skateboard  \\
How long do these animals live?                  & 20 years    \\
What type of train is this?                      & passenger   \\
What is this used for?                           & travel      \\
What is this room used for?                      & sleep       \\
What kind of bird is that?                       & hawk        \\
What food does the animal eat?                   & cat food    \\
What type of dog is this?                        & chihuaha    \\
What food do these animals eat?                  & cat food    \\
What place is this?                              & switzerland \\
What breed of cat is this?                       & calico      \\
What season is this?                             & winter     \\
\bottomrule
\end{tabular}
\caption{Example of random guesses with only question input.
Random guess achieved a good VQA Score by matching to the answers by chance.
But the OK-VQA questions are still not directly answerable without access to the associated images.
}
\label{tab:question_only}
\end{table}
From the feature ablation study we found that our RA-VQA-NoDPR achieved $\sim$28 VQA Score relying on only questions.
This is due to the fact that $\sim75\%$ of answers to training questions appear in the answers to test questions.
As shown in Table~\ref{tab:question_only}, for each distinct question, the model learned to generate the same answer without access to the associated images.
These random guesses can match to the answers of some test questions by chance, leading to a good VQA Score.
By inspection we report that most of the successful cases are random guesses, and these questions are still not directly answerable without reading the associated images.

\section{Case Study}
\label{sec:appendix:case_study}

\begin{figure*}[!bp]
    \centering
    \includegraphics[width=\textwidth]{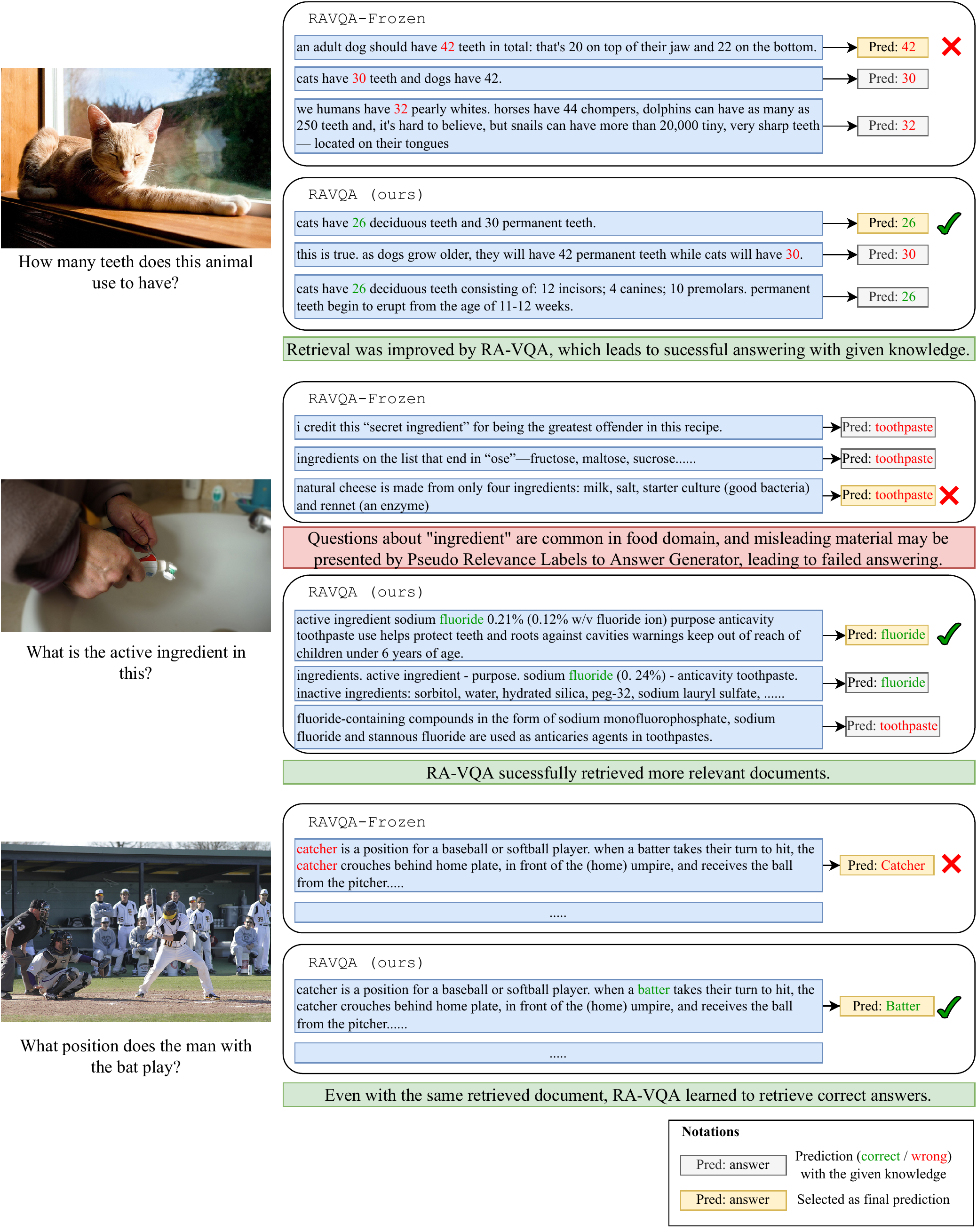}
    \caption{A case study comparing RA-VQA-FrDPR (baseline) and our RA-VQA that benefits from joint training of retrieval and answer generation.}
    \label{fig:case_study}
\end{figure*}

We present a case study in Fig.~\ref{fig:case_study} to compare RA-VQA-FrDPR and our proposed RA-VQA framework.
Conclusions are provided to each case in the figure.

\newpage
\section{Generalizing RAVQA to Other Datasets}
\label{sec:RAVQA:appendix:FVQA_results}
We are also interested in whether this approach is also generalisable to other similar VQA tasks that may benefit from improved passage retrieval.

We implement our framework on another knowledge-based VQA task, Fact-based VQA (FVQA)~\cite{wang2017fvqa}.
This dataset contains commonsense factoid VQA questions, such as ``Question: which object in the image can cut you? Answer: the knife''.
In contrast to OKVQA where no knowledge base is provided, FVQA grounds each question-answer pair with a fact (a triplet from several `common sense' knowledge bases, including ConceptNet~\cite{speer2017conceptnet}, Webchild~\cite{tandon2017webchild}, and DBpedia~\cite{auer2007dbpedia}).
A triplet contains a head node, a relation, and a tail node (e.g. [Car] \texttt{/r/HasA} [4 wheels]).
To cope with passage retrieval, these knowledge triplets are flattened into surface texts (e.g. ``[car] has [4 wheels]'') such that DPR can be directly applied to retrieve them.
We replace pseudo relevance with ground-truth relevance since relevant triplets for answering questions are given.

The metrics used for assessing performance are Accuracy and Recall, with their standard deviations of 5 splits.
Accuracy counts the portion of questions that are successfully answered, while Recall@$K$ measures how likely the retrieved $K$ knowledge triplets contain the answer node.
Since FVQA was designed for answer selection instead of open-ended answer generation, prior works used accuracy as ``whether the answer node is successfully selected from all KG nodes''.
To enable fair evaluation with our open-ended framework, in calculating accuracy, a question is considered successfully answered if the answer node is the closest node to the generated answer string (shortest in Levenshtein distance).
For example, the generated answer `knives' is still a valid answer since the answer node `[knife]' can be matched with a shortest Levenshtein distance.

The significance of performance is guaranteed by reporting the average of 5 splits (as in the official FVQA evaluation).
In total we trained 5 DPR models and 5 $\times$ 3 models (RAVQA, RAVQA-FrDPR, and RAVQA-NoDPR) with the same hyperparameters. Each split has approximately half questions for training and the remaining for testing.

We compare with three systems in prior work:

(1) FVQA~\cite{wang2017fvqa}: the baseline system provided in the official FVQA dataset paper.

(2) GCN~\cite{fvqa_gcn}: a model that leverages graph convolutional networks (GCNs) to aggregate features from visual/language/fact modalities.

(3) Mucko~\cite{zhu2020mucko}: the current state-of-the-art system that uses GCNs to combine visual, fact, and semantic graphs.

\begin{table}[!ht]
\centering
\small
\begin{tabular}{lrrrr}
\toprule
Model       & Accuracy (Std.) & Recall@5 (Std.) \\
\midrule
Mucko       & 73.06 (~~--~~)        & -    \\
\textbf{RAVQA}      & 69.88 (0.13) & 68.77 (0.87) \\
GCN        & 69.35 (~~--~~)         &     -     \\
\textbf{RAVQA-FrDPR} & 68.81 (0.59) & 64.54 (0.80) \\
\textbf{RAVQA-NoDPR} & 67.93 (0.82) & -     \\
FVQA & 58.76 (0.92) & - \\
\bottomrule
\end{tabular}
\caption{Model performance on the FVQA dataset (sorted by accuracy).
Our proposed systems are in bold.}
\label{tab:RAVQA:performance_on_FVQA}
\end{table}

As shown in Table~\ref{tab:RAVQA:performance_on_FVQA}, RAVQA-NoDPR achieves an already strong result (67.93\% accuracy) compared to early work in FVQA, showing that the extracted vision-to-language features are useful and text-based Transformers are able to learn to answer commonsense VQA questions well without accessing the provided knowledge graph (ConceptNet).
The incorporation of DPR boosts the performance to 68.81\% with 64.54\% Recall@5, showing that retrieval works as expected and the retrieved knowledge triplets are exploited in answer generation.
The joint training scheme improves the retrieval (64.54\% to 68.77\% Recall@5) as well as the overall performance (68.81\% to 69.88\% Accuracy).
This demonstrates that our proposed joint training framework is generalizable to other KB-VQA tasks, though the passages used in retrieval are simply flattened surface texts of KG triplets.

In comparing with other systems in the FVQA benchmark, our best system ranks second without an explicit design for leveraging KG structures.
This shows the power of open-ended answer generation with text-based Transformers.
But we emphasise that better performance could be achieved through designing a more specialised retrieval component for the structured knowledge base used in this task.

To summarise, our system shows great generalizability in an external KB-VQA task that was constructed very differently.
Therefore, the proposed framework can serve as a strong basis for future improvements.

\begin{table*}[!ht]
\centering
\small
\begin{tabular}{lcclllccr}
\toprule
Model                   & T5 & GPT-3           & $K_{\text{train}}$ & $K_{\text{test}}$ & Knowl. Src.       & PRRecall  &  EM                                & VQA Score \\
\midrule
 ConceptBERT             &     $\times$ & $\times$      &  -   & -    & C                                         &&      & 33.66      \\
 KRISP                   &     $\times$ & $\times$                       &    -      &  -       & C + W && & 38.35      \\
 VRR & $\times$ & $\times$                        & 100      & 100     & GS (trn/full)                   &&                          & 39.22 / 45.08      \\
 MAVEx                   &     $\times$ & $\times$    & -   &     -    & W + C + GI                  && & 39.40       \\
 KAT-T5                     & \checkmark & $\times$              & 40       & 40      & W                        &&                         & 44.25      \\
TRiG    & \checkmark & $\times$              & 5      & 5     & W                                                && 49.21 & 45.51       \\
TRiG    & \checkmark & $\times$              & 100      & 100     & W                                            &&  53.59  & 49.35       \\
TRiG-Ensemble    & \checkmark & $\times$        & 100      & 100     & W                                        &&    54.73    & 50.50       \\
TRiG*                   & \checkmark & $\times$              & 5        & 5       & GS (trn/full)                            &     &  52.79             & 45.11 / 48.32      \\
RAG*  &   \checkmark & $\times$  & 5 & 5 & GS (trn/full) & 82.34  &  52.52 & 44.90~/~48.22\\
RA-VQA (Ours)                     & \checkmark & $\times$              & 5        & 5       & GS (trn/full)                    & 82.84  & 58.72    & 48.77 / 53.81  \\
RA-VQA (Ours)                     & \checkmark & $\times$              & 5        & 50     & GS (trn/full)    & 96.55                   &                  \textbf{59.41}  & 49.24 / \textbf{54.48}      \\
\midrule
\textit{Ablation Study}\\
\midrule
RA-VQA-FrDPR                    & \checkmark & $\times$              & 5        & 5       & GS (trn/full)  & 81.25 &   55.77   & 47.05 / 51.22      \\
RA-VQA-NoPR                    & \checkmark & $\times$              & 5        & 5       & GS (full)  & 77.67 &   57.80  & 52.98      \\
RA-VQA-FrDPR-NoCT                    & \checkmark & $\times$              & 5        & 5       & GS (full) & 81.25 &   54.99  & 50.66      \\
RA-VQA-NoCT                    & \checkmark & $\times$              & 5        & 5       & GS (full) & 83.77 &  57.51  & 52.67      \\

\midrule
\multicolumn{7}{l}{\textit{GPT-3-based Systems ($>$175 Billion Parameters)}}\\
\midrule
 PICa                    & $\times$ & \checkmark                &     -     & -        & GPT-3                 &&                              & 48.00         \\
KAT-Knowledge-T5            & \checkmark & \checkmark &    40      &   40      & W + GPT-3                                             &&    & 51.97      \\
KAT-Ensemble                     & \checkmark & \checkmark & 40       & 40      & W + GPT-3                                                && & 54.41      \\
\bottomrule
\end{tabular}
\caption{Full table of RA-VQA vs Baseline Systems.
Knowledge Sources: {\bf\underline C}onceptNet; {\bf\underline W}ikipedia; {\bf\underline G}oogle {\bf\underline S}earch; {\bf\underline G}oogle {\bf\underline I}mages; {\bf\underline{GPT-3}} closed book knowledge.   `GS (trn/full)' indicates if the Google Search data is constrained to contain only documents relevant  to training questions.  
PRRecall, HSR, FSR, and EM are reported on  `GS (full)' systems as percentage (\%).
PRRecall is reported at the corresponding $K_{\text{test}}$.
}
\label{tab:comparison_with_baselines_full}
\vspace{-0.5cm}
\end{table*}

\end{document}